\documentclass[11pt]{scaleai-paper}

\usepackage{amsmath}
\usepackage{amsfonts}
\usepackage{amssymb}
\usepackage{booktabs}
\usepackage{multirow}
\usepackage[square,numbers,sort&compress]{natbib}
\usepackage{url}
\usepackage{tikz}
\usetikzlibrary{positioning, arrows.meta, fit, backgrounds, calc, shapes.geometric, shapes.misc}
\usepackage[colorlinks=true,linkcolor=scaleLink,citecolor=scaleLink,urlcolor=scaleLink]{hyperref}
\usepackage[capitalise,nameinlink]{cleveref}

\makeatletter
\providecommand{\@LN@col}[1]{}
\providecommand{\@LN}[2]{}
\makeatother

\papertype{Scale AI Research}
\contact{\texttt{vipul.gupta@scale.com}, \texttt{yunzhong.he@scale.com}}

\title{CRAFT: Clustering Rubrics to Diagnose Weak LLM Capabilities and Generate Targeted Fine-Tuning Data}

\author[1]{Vipul Gupta}
\author[1]{Zihao Wang}
\author[1]{Razvan-Gabriel Dumitru}
\author[1]{MohammadHossein Rezaei}
\author[1]{Aakash Sabharwal}
\author[1]{Yunzhong He}
\affil[1]{Scale AI}

\hypersetup{
  pdftitle={CRAFT: Clustering Rubrics to Diagnose Weak LLM Capabilities and Generate Targeted Fine-Tuning Data},
  pdfauthor={Vipul Gupta, Zihao Wang, Razvan-Gabriel Dumitru, MohammadHossein Rezaei, Aakash Sabharwal, Yunzhong He}
}

\begin{document}
\maketitle
\begin{abstract}
Evaluations should do more than measure a model's current performance. They should tell us what to fix for the next model iteration and provide a way to generate targeted post-training data. Most evaluation pipelines identify weak examples, topics, or categories, but they leave the underlying capability failure implicit: they say \emph{where} a model fails, not \emph{why}. We introduce CRAFT, a method that converts any rubric-based evaluation dataset into a model-specific diagnosis of weak capabilities. CRAFT treats each grading criterion as a capability probe: it extracts a capability description from every prompt--rubric pair, clusters these descriptions into a hierarchical capability tree, scores the target model at every node, and selects low-performing nodes dynamically across tree levels, at the granularity where each failure is clearest. The selected weak capabilities then direct the generation of targeted supervised fine-tuning data. Holding the data-generation, fine-tuning, and evaluation setup fixed, we compare CRAFT against prompt-level EvalTree clustering and untargeted random generation on four open-source models, two professional domains (finance and legal), and 13 held-out benchmarks disjoint from the diagnostic data. CRAFT achieves the strongest finance-domain average for all four models under repeated temperature decoding; on legal domain, it is strongest for three of four models and remains within the decoding-variance bands of the best baseline on the fourth. Diagnosing weaknesses at the level of rubric criteria, rather than prompts or categories, thus yields both a sharper picture of what a model cannot do and measurably better models after fine-tuning on that diagnosis.
\end{abstract}

\section{Introduction}

Evaluations of large language models serve two purposes. The first, measuring how a model performs today, has received enormous attention: broad suites and targeted benchmarks rank models across ever more scenarios, tasks, and metrics \cite{hendrycks2021measuring, rein2024gpqa, liang2023holistic, srivastava2023beyond}. The second, telling us what to fix so the model performs better tomorrow, remains underdeveloped, even though it is the purpose that matters most for post-training data collection. Benchmark results are still typically summarized as task- or category-level scores. Such summaries identify \emph{where} a model is weak; they rarely identify the failed answer requirements precisely enough to guide the next fine-tuning data collection step.

This gap matters because a single weak score conflates many distinct failures. Capability- and behavior-oriented evaluations \cite{ribeiro2020beyond, ye2024flask} move toward finer resolution by decomposing model behavior into specific skills or tests. But even under a skill-oriented view, a low score on a math benchmark may reflect choosing the wrong equation, making an arithmetic error, skipping an intermediate step, or omitting the required units, failures that call for different interventions even when they occur on the same prompt. An evaluation that stops at ``weak on math'' cannot choose among them.

Rubrics provide a natural substrate for this finer diagnosis. Rubric criteria decompose answer quality into explicit requirements: for example, a math-problem rubric might separately check whether the solution identifies the relevant quantities, sets up the correct equation, solves it accurately, and states the final answer with the right units. Rubrics are increasingly used in evaluation and reinforcement learning \cite{rezaei2025online, viswanathan2026checklists, gunjal2025rubrics, liu-etal-2023-g, kim2024prometheus, wu2023fine}, but existing work treats criteria as scoring devices: to make judgments or to train reward models. We instead treat a rubric dataset as a pool of thousands of capability probes. Each criterion measures something the model must be able to do. When aggregated across a dataset, rubrics reveal recurring weaknesses that are invisible when each rubric is treated individually.

We propose CRAFT (Clustering Rubrics for Actionable Fine-Tuning), a method that converts rubric-annotated evaluation data into a model-specific diagnosis of weak capabilities. CRAFT extracts a capability description from each prompt--rubric pair, clusters these descriptions into a hierarchical capability tree, and evaluates the target model at every node. Because weaknesses do not always appear at the same granularity, CRAFT selects low-performing nodes dynamically across tree levels rather than committing to a fixed depth. After CRAFT has produced this weak-node set, a separate downstream generation step uses the selected nodes to produce fine-tuning data focused on the model's weak capabilities.

We evaluate CRAFT on four open-source models (Qwen3-4B, Qwen3-8B, Gemma-3-4B, and Llama-3.1-8B-Instruct) in two professional domains, finance and legal. We compare against EvalTree \cite{zeng2025evaltree}, which builds a capability tree over whole prompts rather than rubric criteria, and against a Random baseline that runs a non-tree-based random data selection approach without weakness targeting; all methods share the same downstream generation, fine-tuning, and evaluation setting, so they differ only in how training data is targeted. On 13 held-out benchmarks, seven legal and six finance, all disjoint from the rubric dataset used for diagnosis, CRAFT achieves the best repeated-decoding finance-domain average for all four models and the best repeated-decoding legal performance for three of four models; on the remaining legal model, its score overlaps the decoding-variance band of the best baseline. The pattern highlights the efficacy of our approach: outperforming Random shows the value of building and targeting a scored capability hierarchy at all, while outperforming EvalTree shows that rubric criteria, not prompts, are the more informative unit to organize. We show that under this shared setting, rubric-level diagnosis identifies better targets for downstream fine-tuning data than prompt-level or untargeted selection.

Our main contributions are:
\begin{enumerate}
    \item We introduce CRAFT, a method that clusters a rubric dataset into a hierarchical, model-specific diagnosis of weak capabilities that directly helps in fine-tuning data generation.
    \item Under an identical synthetic-data generation and fine-tuning setting, CRAFT achieves the strongest performance for all four models in the finance domain and strongest performance for three of four tested models in the legal domain.
    \item We show that the granularity of weakness selection matters: dynamic cross-level selection outperforms fixed-level selection, and the better fixed level varies by model and domain, so no single depth suffices.
\end{enumerate}


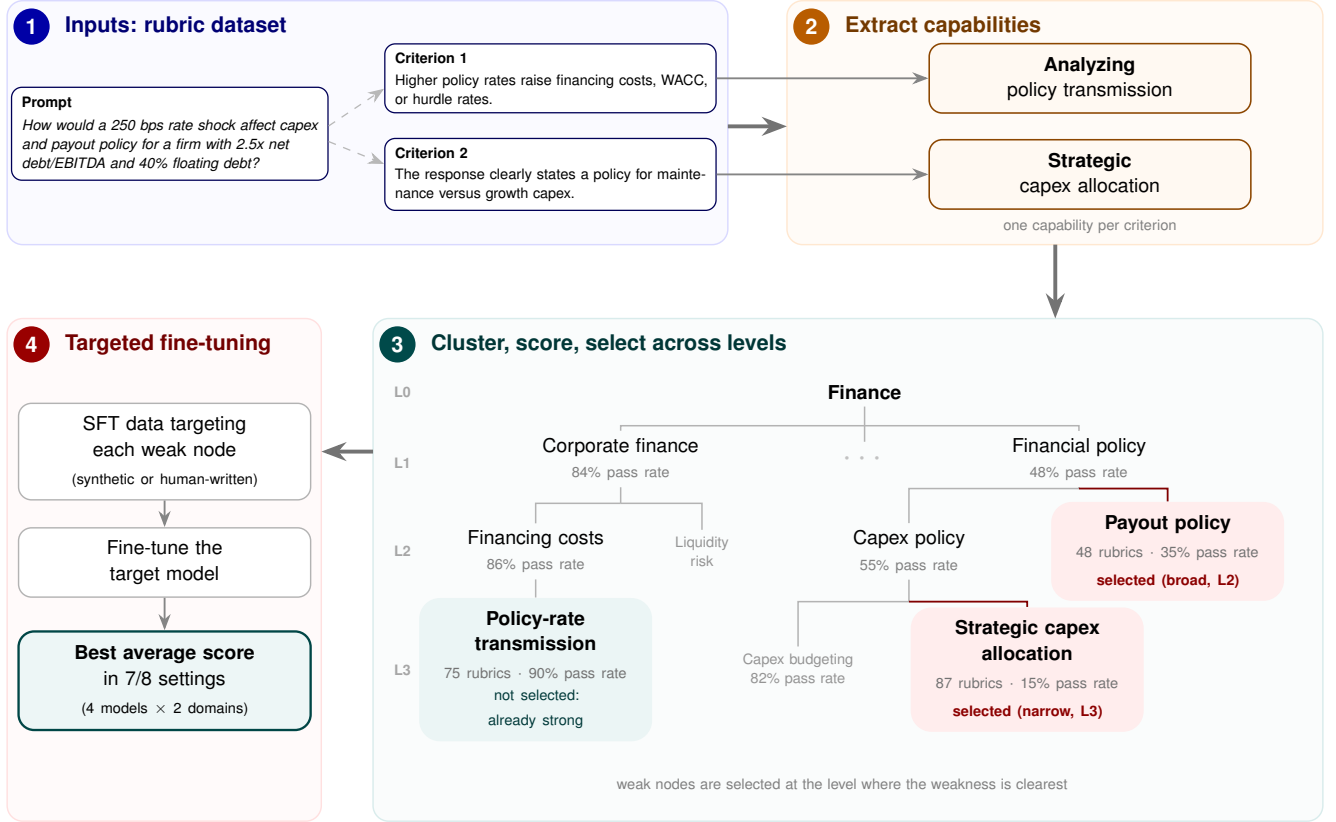
\begin{figure}[!t]
\centering
\resizebox{\textwidth}{!}{%
\begin{tikzpicture}[
  font=\sffamily,
  >={Stealth[length=2.2mm]},
  panelA/.style={rounded corners=6pt, fill=blue!2, draw=blue!15, line width=0.55pt},
  panelB/.style={rounded corners=6pt, fill=orange!3, draw=orange!18, line width=0.55pt},
  panelC/.style={rounded corners=6pt, fill=teal!2, draw=teal!15, line width=0.55pt},
  panelD/.style={rounded corners=6pt, fill=red!2, draw=red!12, line width=0.55pt},
  badge/.style={circle, text=white, font=\sffamily\bfseries\footnotesize, inner sep=0pt, minimum size=4.9mm},
  hdr/.style={font=\sffamily\bfseries\footnotesize, anchor=west, align=left},
  smallnote/.style={font=\sffamily\tiny, text=black!50, align=center},
  pcard/.style={rounded corners=4pt, draw=blue!34!black, fill=white, font=\sffamily\tiny, align=left, inner xsep=4pt, inner ysep=3.5pt, text width=40mm, line width=0.5pt},
  capbox/.style={rounded corners=5pt, draw=orange!55!black, fill=orange!5, font=\sffamily\scriptsize, align=center, inner sep=5pt, text width=40mm, line width=0.6pt},
  tnode/.style={font=\sffamily\scriptsize, align=center, text width=#1, inner xsep=1pt, inner ysep=2.5pt},
  tnode/.default=24mm,
  faintnode/.style={font=\sffamily\tiny, align=center, text=black!42, text width=#1, inner xsep=1pt, inner ysep=2.5pt},
  faintnode/.default=18mm,
  weak/.style={rounded corners=8pt, fill=red!7, font=\sffamily\scriptsize, align=center, inner xsep=5pt, inner ysep=5pt, text width=#1},
  weak/.default=28mm,
  strong/.style={rounded corners=8pt, fill=teal!8, font=\sffamily\scriptsize, align=center, inner xsep=5pt, inner ysep=5pt, text width=28mm},
  chip/.style={rounded corners=5pt, draw=black!30, fill=white, font=\sffamily\scriptsize, align=center, inner xsep=5pt, inner ysep=5pt, text width=36mm, line width=0.55pt},
  outcome/.style={rounded corners=6pt, draw=teal!55!black, fill=teal!7, font=\sffamily\scriptsize, align=center, inner xsep=5pt, inner ysep=5.5pt, text width=36mm, line width=0.9pt},
  lvl/.style={font=\sffamily\tiny\bfseries, text=black!35},
  arr/.style={->, draw=black!48, line width=0.72pt},
  arrRed/.style={->, draw=red!55!black, line width=0.8pt},
  flow/.style={->, >={Stealth[length=3.4mm]}, draw=black!55, line width=1.5pt},
  flowRed/.style={->, >={Stealth[length=3.4mm]}, draw=red!55!black, line width=1.5pt},
  pairline/.style={->, draw=black!28, line width=0.5pt, dashed},
  edge/.style={draw=black!30, line width=0.58pt},
  edgeRed/.style={draw=red!50!black, line width=0.7pt},
]

\draw[panelA] (0.15,8.55) rectangle (9.90,11.85);
\draw[panelB] (10.70,8.55) rectangle (17.95,11.85);
\draw[panelC] (5.10,0.75)  rectangle (17.95,7.55);
\draw[panelD] (0.15,0.75) rectangle (4.40,7.55);

\node[badge, fill=blue!65!black]  at (0.48,11.50){1}; \node[hdr, text=blue!45!black]   at (0.80,11.50){Inputs: rubric dataset};
\node[badge, fill=orange!72!black] at (11.03,11.50){2}; \node[hdr, text=orange!45!black] at (11.35,11.50){Extract capabilities};
\node[badge, fill=teal!58!black]  at (5.43,7.20){3};  \node[hdr, text=teal!42!black]   at (5.75,7.20){Cluster, score, select across levels};
\node[badge, fill=red!60!black]   at (0.48,7.20){4}; \node[hdr, text=red!45!black]    at (0.80,7.20){Targeted fine-tuning};

\draw[flow] (9.90,10.15) -- (10.70,10.15);        
\draw[flow] (14.33,8.55) -- (14.33,7.55);         
\draw[flow] (5.10,5.75) -- (4.40,5.75);           

\node[pcard] (prm) at (2.35,10.05)
  {\textbf{Prompt}\\[2pt]\emph{How would a 250 bps rate shock affect capex and payout policy for a firm with 2.5x net debt/EBITDA and 40\% floating debt?}};
\node[pcard, text width=42mm] (rc1) at (7.50,10.80)
  {\textbf{Criterion 1}\\[1pt]Higher policy rates raise financing costs, WACC, or hurdle rates.};
\node[pcard, text width=42mm] (rc2) at (7.50,9.50)
  {\textbf{Criterion 2}\\[1pt]The response clearly states a policy for maintenance versus growth capex.};

\node[capbox] (h1) at (14.80,10.80) {\textbf{Analyzing}\\ policy transmission};
\node[capbox] (h2) at (14.80,9.50)  {\textbf{Strategic}\\ capex allocation};
\draw[arr] (rc1.east) -- (h1.west);
\draw[arr] (rc2.east) -- (h2.west);
\draw[pairline] ([yshift=3pt]prm.east)  -- ([yshift=-4pt]rc1.west);
\draw[pairline] ([yshift=-3pt]prm.east) -- ([yshift=4pt]rc2.west);
\node[smallnote, text width=40mm] at (14.80,8.80)
  {one capability per criterion};

\node[lvl] at (5.50,6.55) {L0};
\node[lvl] at (5.50,5.60) {L1};
\node[lvl] at (5.50,4.40) {L2};
\node[lvl] at (5.50,2.80) {L3};

\node[tnode=24mm] (root) at (11.75,6.55)
  {\textbf{Finance}};

\node[tnode=27mm] (fa) at (8.45,5.65)
  {Corporate finance\\[0pt]{\tiny\textcolor{black!50}{84\% pass rate}}};
\node[font=\large, text=black!24] (cmid) at (11.75,5.65) {$\cdots$};
\node[tnode=27mm] (pu) at (14.65,5.65)
  {Financial policy\\[0pt]{\tiny\textcolor{black!50}{48\% pass rate}}};

\node[tnode=22mm] (costs) at (7.30,4.40)
  {Financing costs\\[0pt]{\tiny\textcolor{black!50}{86\% pass rate}}};
\node[faintnode=16mm] (risk) at (9.55,4.40) {Liquidity\\ risk};
\node[tnode=22mm] (capex) at (12.35,4.40)
  {Capex policy\\[0pt]{\tiny\textcolor{black!50}{55\% pass rate}}};
\node[weak] (payout) at (15.85,4.40)
  {\textbf{Payout policy}\\[1pt]{\tiny\textcolor{black!55}{48 rubrics $\cdot$ 35\% pass rate}}\\[1pt]{\tiny\bfseries\textcolor{red!55!black}{selected (broad, L2)}}};

\node[strong] (n1) at (7.30,2.80)
  {\textbf{Policy-rate transmission}\\[1pt]{\tiny\textcolor{black!55}{75 rubrics $\cdot$ 90\% pass rate}}\\[1pt]{\tiny\textcolor{teal!45!black}{not selected:\\ already strong}}};
\node[faintnode=17mm] (budg) at (10.85,2.80)
  {Capex budgeting\\{\tiny 82\% pass rate}};
\node[weak] (n2) at (13.95,2.80)
  {\textbf{Strategic capex allocation}\\[1pt]{\tiny\textcolor{black!55}{87 rubrics $\cdot$ 15\% pass rate}}\\[1pt]{\tiny\bfseries\textcolor{red!55!black}{selected (narrow, L3)}}};

\node[smallnote, text width=110mm] at (11.45,1.25)
  {weak nodes are selected at the level where the weakness is clearest};

\coordinate (rjoin) at (11.75,6.10);
\coordinate (fjoin) at (8.45,5.05);
\coordinate (pjoin) at (14.65,5.25);
\coordinate (cjoin) at (12.35,3.72);
\draw[edge] (root.south) -- (rjoin);
\draw[edge] (rjoin) -| (fa.north);
\draw[edge] (rjoin) -- (cmid.north);
\draw[edge] (rjoin) -| (pu.north);
\draw[edge] (fa.south) -- (fjoin);
\draw[edge] (fjoin) -| (costs.north);
\draw[edge] (fjoin) -| (risk.north);
\draw[edge] (pu.south) -- (pjoin);
\draw[edge] (pjoin) -| (capex.north);
\draw[edgeRed] (pjoin) -| (payout.north);
\draw[edge] (costs.south) -- (n1.north);
\draw[edge] (capex.south) -- (cjoin);
\draw[edge] (cjoin) -| (budg.north);
\draw[edgeRed] (cjoin) -| (n2.north);

\node[chip] (d1) at (2.28,5.75)
  {SFT data targeting\\ each weak node\\[1pt]{\tiny (synthetic or human-written)}};
\node[chip] (d2) at (2.28,4.25)
  {Fine-tune the\\ target model};
\node[outcome] (d3) at (2.28,2.65)
  {\textbf{Best average score}\\ in 7/8 settings\\[1pt]{\tiny (4 models $\times$ 2 domains)}};
\draw[arr] (d1.south) -- (d2.north);
\draw[arr] (d2.south) -- (d3.north);

\end{tikzpicture}%
}
\caption{\textbf{CRAFT} pipeline. CRAFT extracts per-criterion capabilities, builds and scores a hierarchy, selects weak nodes (red), and uses them to generate targeted fine-tuning data.}
\label{fig:pipeline}
\end{figure}

\section{Related Work}

\subsection{Rubrics}

Rubric-based evaluation decomposes a response judgment into explicit criteria rather than asking for a single holistic score. This makes the grading signal more interpretable: a model can be right on factual grounding but wrong on completeness, or correct in reasoning but wrong in final-answer format. Benchmarks such as HealthBench \cite{arora2025healthbench} and PRBench \cite{akyurek2025prbench} use large collections of expert-written rubrics in professional domains, and related datasets extend rubric supervision to other specialized and post-training settings \cite{zhao2026pancanbench, li2026rubrichub}. A second line of work uses rubric-like criteria to improve LLM-based evaluation, where criteria make the judge's decision process more explicit and fine-grained \cite{liu-etal-2023-g, kim2024prometheus}. Another line of work uses criteria or fine-grained feedback as post-training signals, including reward modeling and AI-feedback pipelines \cite{zhang2025chasing, rezaei2025online, wu2023fine, mahmoud2026reward, gunjal2025rubrics, cui2024ultrafeedback}. CRAFT uses the same criterion-level signal for a different purpose: instead of treating criteria only as independent scoring items or rewards, it clusters them across the dataset to recover a structured map of recurring capabilities and failures.

\subsection{Prompt-Level Hierarchical Clustering and EvalTree}

The closest prior method is EvalTree~\cite{zeng2025evaltree}, which profiles a model by organizing evaluation prompts into a hierarchical capability tree. EvalTree clusters whole prompts, describes each internal node with a natural-language capability label, scores the target model on each subtree, and uses low-performing prompt clusters to guide targeted data generation. A prompt-level node can say that a model struggles with a kind of question; it cannot isolate which capability inside the answer was missed. CRAFT clusters prompt--rubric pairs instead. This lets one prompt contribute multiple leaves, one for each criterion it tests, and lets a node represent a response capability such as applying a statutory exception or grounding a financial calculation in the correct line item. The comparison to EvalTree therefore directly tests whether rubric criteria add diagnostic information beyond prompt clustering when the selected data are passed through the same downstream fine-tuning workflow.

\section{Methodology}

CRAFT takes as input (i) a rubric dataset, i.e., prompts annotated with one or more grading criteria as illustrated in Figure~\ref{fig:pipeline}, and (ii) a target language model. It returns a model-specific set of weak capability nodes intended to serve as fine-tuning targets, not merely as a finer-grained score report. The methodology converts rubric criteria into capability descriptions, hierarchically clusters those capabilities, scores the target model at every node, and selects low-performing nodes. Section~\ref{sec:fine_tuning_generation} describes the separate downstream step that uses those selected nodes to generate training examples.

\subsection{Rubrics}

A rubric is a set of explicit requirements that a response must satisfy. Each criterion is a short natural-language statement, such as ``the response cites the controlling statute'' or ``the response identifies the cash-flow line item before computing the ratio.'' Each criterion can be scored independently as pass/fail or with a small discrete score. A response might pass some criteria and fail others. This makes rubrics useful not only as grading instruments, but also as capability probes: each criterion names something the model must be able to do for the answer to be acceptable. In this sense, a criterion resembles a small behavioral test of one required skill, connecting rubric scoring to fine-grained behavioral evaluation \cite{ribeiro2020beyond, ye2024flask}. CRAFT builds on this observation by aggregating criteria across a dataset rather than using them only as per-example scoring rules.

\subsection{Extracting Capabilities from Rubrics}

A single prompt can contain many criteria, and those criteria may test different skills. We therefore flatten the dataset into prompt--rubric pairs and process each pair independently. For every pair, a separate language model produces a short, decontextualized description of the capability needed to satisfy the criterion. The output is a collection of tuples containing the original prompt, the rubric criterion, and the extracted capability description. This representation allows criteria to cluster together even when they come from different prompts or use different wording, as long as they test the same underlying capability.

\subsection{Hierarchical Clustering}

We organize the extracted capability descriptions into a tree with a bottom-up, level-by-level procedure adapted from the hierarchizer of Clio~\cite{tamkin2024clio}. Unlike classical agglomerative clustering, which greedily merges the closest pair of nodes at each step, our method constructs one entire level at a time: given the clusters at the current level, an LLM writes a smaller inventory of more general capability labels, every current cluster is routed to exactly one label, and the process repeats until the level contains a target number of top clusters. Leaves correspond to individual prompt--rubric pairs, and every internal node carries a natural-language capability description that is more general than its children while preserving their domain content (Figure~\ref{fig:pipeline}).

\paragraph{Fixed top levels.} When the rubric dataset already provides categorical metadata, such as a domain label or a per-criterion subcategory annotation, we use it to define the uppermost levels of the tree rather than learning them: each metadata attribute becomes one fixed level (e.g., domain at level~0, annotated subcategory at level~1). Capabilities are then clustered independently within each resulting group, and the learned subtrees are grafted under the corresponding fixed nodes. This guarantees that learned clusters never mix annotated categories and preserves any taxonomy the dataset authors intended. If no such metadata is available, the entire dataset forms a single group and the full tree is learned.

\paragraph{Base-level clusters.} Within each group, we embed every capability description and partition the embeddings with $k$-means into $n_{\mathrm{base}}$ base clusters, the finest level of the learned hierarchy. Both $n_{\mathrm{base}}$ and the target number of top clusters $n_{\mathrm{top}}$ are set adaptively from the group size, so larger groups receive proportionally finer base partitions. An LLM then names each base cluster from a sample of its member capabilities, producing the capability description used at all later stages.

\paragraph{Building higher levels.} The depth of the tree is not a hyperparameter fixed in advance; it is determined dynamically from the data. The number of learned levels is set to $\max\!\big(2, \lfloor \log_2 (n_{\mathrm{base}} / n_{\mathrm{top}}) \rfloor + 1\big)$, and because $n_{\mathrm{base}}$ and $n_{\mathrm{top}}$ are themselves derived from the group size, larger rubric collections automatically produce deeper hierarchies while small ones stay shallow. Intermediate level sizes are interpolated geometrically between $n_{\mathrm{base}}$ and $n_{\mathrm{top}}$, and construction stops early if a level already reaches the top-cluster target, so the realized depth can also differ across groups within one dataset.

Each higher level is built from the one below it in four steps. (i)~\emph{Node summarization}: the child clusters are embedded by name and description and grouped with $k$-means into neighborhoods of ${\sim}40$ clusters; for each neighborhood, an LLM reads the in-neighborhood clusters, plus a few out-of-neighborhood clusters for contrast, and writes candidate parent capabilities near the level's target count. (ii)~\emph{Candidate consolidation}: candidate capabilities from all neighborhoods are pooled, and a second LLM pass combines duplicate, overlapping, or unnecessarily fine candidates into one parent set. (iii)~\emph{Cluster routing}: an LLM routes each child cluster to the best-fitting parent, with parent order shuffled to reduce position bias; a child's prompt--rubric pairs propagate to its parent, so every node aggregates the pairs in its subtree. (iv)~\emph{Parent refinement}: each parent is re-described from the children routed to it, so node descriptions reflect final tree contents rather than the initial candidate text. During node summarization, the LLM can also invoke a validation tool that repeatedly routes a sample of child clusters to the candidate parents with an independent model; children that flip between parents across repeated trials indicate overlapping candidates, and the candidate set is revised until routing is stable.

In our experiments, the datasets provide domain and rubric-subcategory annotations, so these occupy the two fixed top levels and the procedure yields a six-level tree (L0 domain root to L5 finest) over 20{,}443 prompt--rubric pairs. We use Gemini~2.5~Pro \cite{comanici2025gemini} for node summarization, candidate consolidation, cluster routing, and parent refinement, Gemini~2.5~Flash \cite{comanici2025gemini} for the routing-stability checks, and \texttt{text-embedding-3-small} \cite{openai_text_embedding_3_small} for all embeddings.

The hierarchy is important because different failures appear at different depths. A broad node may capture a general weakness, such as reasoning over regulatory obligations, while a child node may isolate a narrower failure, such as applying anti-avoidance rules in a cross-border tax setting. A fixed-depth analysis can therefore either fragment broad weaknesses or merge narrow weaknesses into overly general categories.

\subsection{Dynamically Identifying Low-Performing Nodes}

To localize the capabilities on which the target model underperforms, we first run the model on every prompt in the rubric dataset and score each response against its associated criteria with an LLM-as-a-judge \cite{liu-etal-2023-g, kim2024prometheus, zeng2024evaluating}. This yields a pass/fail label for each prompt--rubric pair, which corresponds to a leaf in the capability tree. This judge step is important because all node scores inherit the criterion-level judge decisions; CRAFT therefore depends on judges that can apply rubric criteria consistently rather than only produce a holistic preference. We then aggregate these leaf scores upward: each internal node receives the average pass rate over the prompt--rubric pairs in its subtree. As a result, every node has both a semantic description and an empirical estimate of how reliably the target model satisfies that capability.

This scored tree lets CRAFT choose weak capabilities at the granularity where they are most clearly expressed. In our experiments, different models often exhibit weaknesses at different depths: a stronger model may fail only on narrow descendant nodes, whereas a weaker model may struggle with broader parent capabilities closer to the root. CRAFT therefore selects low-performing nodes with a top-down cross-level search rather than committing to a fixed tree depth. A node is eligible only if it has sufficient support--more than 30 associated rubric criteria in our main experiments--so selected nodes reflect stable capability estimates rather than sparse artifacts. The search starts with a strict pass-rate threshold and relaxes it incrementally until the target number of weak nodes is collected. Once a node is selected, its ancestors and descendants are excluded, preventing overlapping selections and leaving a set of disjoint weak-capability regions. This procedure is intentionally simple: it uses the tree to choose a useful granularity for each weakness rather than imposing one global depth. Section~\ref{sec:ablation} tests this design against fixed-level alternatives.

The output of CRAFT is a list of the target model's weak capability nodes, each grounded in a node description, a pass-rate estimate, and supporting prompt--rubric pairs. The downstream fine-tuning data-generation step uses these nodes as targeting information.

\section{Fine-Tuning Data Generation}
\label{sec:fine_tuning_generation}

After the CRAFT methodology has identified weak capabilities in a model, a separate fine-tuning data-generation step turns those measured weaknesses into supervised training examples. These examples could be written by humans or generated synthetically. In our experiments, we use synthetic data generation so that the data-selection strategy can be compared at a fixed scale. This is where the evaluation output becomes an interface for data generation: examples are conditioned on diagnosed weak capabilities rather than on the domain alone. This follows the broader idea that training data can be selected to target particular skills or downstream behaviors, but CRAFT derives those targets from empirical rubric failures rather than from a predefined skill taxonomy \cite{chen2023skill, xia_less_selecting2024}.

For each identified weak capability node, we generate 40 prompts. The generation prompt includes the node description, examples from the node, and nearby tree context such as parent and sibling descriptions. This context directs the generator toward the selected capability while discouraging overlap with adjacent capabilities. We then apply lightweight filtering, including near-duplicate removal with word-overlap Jaccard similarity and diversity constraints on repeated prompt patterns.

Responses are generated by a teacher model using the same response-style recipe for all four models and all three data-selection methods. After generation, each method is trained on the resulting prompt--response pairs with the same supervised fine-tuning setting: a 1{,}000-example target budget, the same training hyperparameters, and the same evaluation protocol. Thus CRAFT, EvalTree, and Random differ in how they select or target synthetic data, not in the downstream fine-tuning setting.

\section{Baseline Methods and Models Used}

\subsection{EvalTree}
We compare against EvalTree~\cite{zeng2025evaltree}, a prompt-level weakness-profiling method. EvalTree organizes prompts into a hierarchical capability tree, scores the model on prompt subtrees, and selects low-performing prompt clusters for data generation. We run the same downstream synthetic-data and fine-tuning pipeline after selection. The difference is therefore the diagnostic unit: whole prompts for EvalTree versus prompt--rubric criteria for CRAFT.

\subsection{Random Baseline}
The Random baseline removes both the capability hierarchy and weakness targeting. It treats the domain pool as a flat, unstructured collection: seed prompts are sampled uniformly at random, without building a capability tree, scoring nodes, or conditioning on diagnosed weaknesses, and are then passed through the same teacher generation and fine-tuning pipeline. This controls for exposure to additional synthetic in-domain data. Because Random differs from CRAFT only in replacing the scored capability hierarchy with flat uniform sampling, CRAFT's improvements over Random measure the value of organizing the domain into a hierarchy and targeting its weak nodes, rather than the value of more data from the same domain. The comparison connects CRAFT to broader work on skill- and influence-guided data selection for language-model training \cite{chen2023skill, xia_less_selecting2024}.

\subsection{Models Used}

We evaluate four open-source models: Qwen3-4B, Qwen3-8B, Gemma-3-4B, and Llama-3.1-8B-Instruct \cite{yang2025qwen3, Kamath2025Gemma3T, grattafiori2024llama}. Each model is evaluated in two domains: finance and legal. For all four models and all three data-selection methods, CRAFT, EvalTree, and Random use the same data budget, teacher-generation procedure, supervised fine-tuning hyperparameters, and evaluation protocol. The diagnostic scoring is model-conditioned, so the selected weak nodes and generated examples can differ across methods, but the experimental setting is shared.

\subsection{Dataset Selection}
We run all experiments in two professional domains: finance and legal. To avoid evaluation contamination, the rubric datasets used for diagnosis are kept separate from the held-out benchmarks used for final reporting. For finance, we use the finance subset of PRBench~\cite{akyurek2025prbench}, which contains 629 prompts and 10{,}806 rubric criteria. For legal, we use the legal subset of PRBench~\cite{akyurek2025prbench}, which contains 532 prompts and 9{,}637 rubric criteria. These subsets are used only to build the capability tree, score model performance on rubric criteria, and select weak capability nodes; they are not used as benchmark test sets.

We choose PRBench because it provides domain-specialized prompts paired with explicit rubrics, which is the supervision CRAFT requires. The held-out benchmark suites used to measure final performance are described in Section~\ref{sec:benchmarks}.

\section{Results}

\begin{figure}[t]
    \centering
    \includegraphics[width=\textwidth]{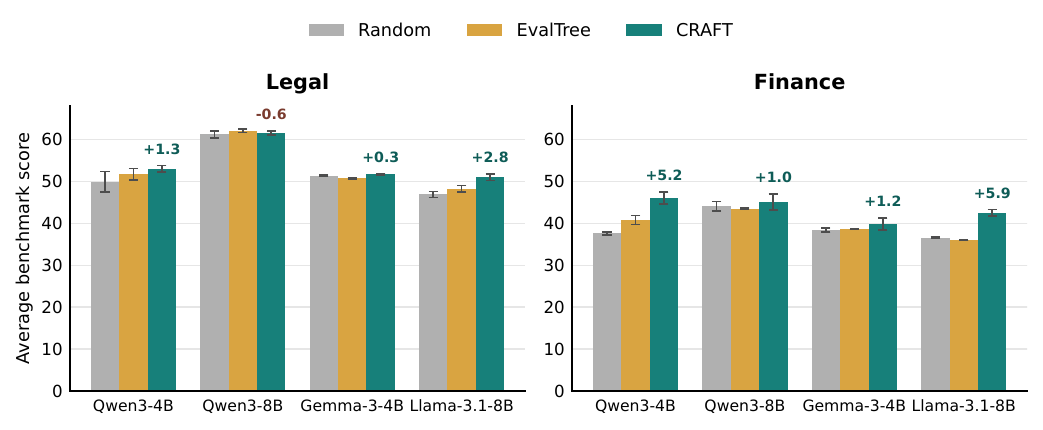}
    \caption{Repeated-decoding average scores from Tables~\ref{tab:legal_craft_wins} and~\ref{tab:finance_craft_wins}. CRAFT (teal) leads finance for all four models and legal for three of four.}
    \label{fig:average_scores}
\end{figure}

We evaluate CRAFT against EvalTree and Random on finance and legal benchmarks chosen to test both domain knowledge and task-style transfer. For every model and data-selection method, the recipe, synthetic-data budget, supervised fine-tuning setup, and scoring protocol are exactly the same. The methods differ only in how they target the fine-tuning data. To reduce noise in final results, we evaluate each trained model--method checkpoint with multiple repeated temperature-decoding runs using distinct sampling seeds. Each cell in Tables~\ref{tab:legal_craft_wins} and~\ref{tab:finance_craft_wins} reports the mean $\pm$ sample standard deviation across those repeated runs for one held-out benchmark. The \textit{Average} row gives the mean $\pm$ sample standard deviation of the unweighted domain average, computed across complete repeated runs. We refer to this mean $\pm$ one-standard-deviation interval as a decoding-variance band. Figure~\ref{fig:average_scores} visualizes the repeated-decoding average rows for both domains.

  \begin{table*}[t]
    \centering
    \caption{Accuracy on legal benchmarks for Random, EvalTree, and CRAFT. Cells report mean $\pm$ standard deviation over repeated temperature-0.5 decoding runs; \textit{Average} is the unweighted benchmark mean. CRAFT performs best in 3 of the 4 models and is within standard deviation in the fourth.}
    \label{tab:legal_craft_wins}
    \small
    \begin{tabular}{l l c c c}
    \toprule
    Model & Benchmark & Random & EvalTree & CRAFT \\
    \midrule
    \multirow{8}{*}{Qwen3-4B}
     & LegalBench         & 52.5\,$\pm$\,5.1 & 60.3\,$\pm$\,5.7 & 66.1\,$\pm$\,0.6 \\
     & CaseHOLD           & 59.9\,$\pm$\,2.2 & 46.3\,$\pm$\,15.9 & 59.8\,$\pm$\,2.0 \\
     & SARA               & 22.9\,$\pm$\,7.1 & 24.1\,$\pm$\,5.8 & 25.4\,$\pm$\,4.1 \\
     & MMLU Law           & 33.3\,$\pm$\,3.5 & 42.4\,$\pm$\,5.1 & 34.9\,$\pm$\,5.5 \\
     & MBE                & 35.3\,$\pm$\,0.4 & 33.3\,$\pm$\,7.6 & 35.0\,$\pm$\,3.4 \\
     & ContractNLI        & 65.4\,$\pm$\,4.3 & 65.9\,$\pm$\,4.7 & 64.3\,$\pm$\,0.5 \\
     & Consumer QA        & 80.2\,$\pm$\,3.9 & 89.7\,$\pm$\,0.5 & 85.5\,$\pm$\,1.1 \\
     & \textit{Average}   & 49.9\,$\pm$\,2.5 & 51.7\,$\pm$\,1.4 & \textbf{53.0\,$\pm$\,0.8} \\
    \cmidrule(lr){1-5}
    \multirow{8}{*}{Qwen3-8B}
     & LegalBench         & 61.0\,$\pm$\,3.4 & 66.5\,$\pm$\,0.7 & 61.5\,$\pm$\,2.3 \\
     & CaseHOLD           & 65.5\,$\pm$\,0.5 & 64.5\,$\pm$\,1.4 & 65.6\,$\pm$\,0.5 \\
     & SARA               & 35.6\,$\pm$\,0.2 & 35.4\,$\pm$\,0.1 & 35.4\,$\pm$\,0.3 \\
     & MMLU Law           & 51.9\,$\pm$\,0.9 & 51.4\,$\pm$\,0.4 & 51.5\,$\pm$\,0.5 \\
     & MBE                & 47.7\,$\pm$\,0.8 & 49.5\,$\pm$\,1.4 & 48.7\,$\pm$\,0.4 \\
     & ContractNLI        & 75.7\,$\pm$\,1.2 & 76.1\,$\pm$\,0.3 & 76.6\,$\pm$\,0.5 \\
     & Consumer QA        & 90.9\,$\pm$\,0.9 & 91.1\,$\pm$\,0.9 & 91.1\,$\pm$\,0.7 \\
     & \textit{Average}   & 61.2\,$\pm$\,0.8 & \textbf{62.1\,$\pm$\,0.4} & 61.5\,$\pm$\,0.5 \\
    \cmidrule(lr){1-5}
    \multirow{8}{*}{Gemma-3-4B}
     & LegalBench         & 65.6\,$\pm$\,0.4 & 64.6\,$\pm$\,0.4 & 66.9\,$\pm$\,0.2 \\
     & CaseHOLD           & 49.7\,$\pm$\,0.7 & 54.0\,$\pm$\,1.0 & 53.0\,$\pm$\,0.8 \\
     & SARA               & 29.5\,$\pm$\,0.7 & 31.6\,$\pm$\,1.0 & 30.9\,$\pm$\,1.1 \\
     & MMLU Law           & 40.9\,$\pm$\,0.5 & 41.9\,$\pm$\,0.6 & 40.7\,$\pm$\,0.2 \\
     & MBE                & 23.8\,$\pm$\,1.3 & 15.8\,$\pm$\,1.8 & 24.2\,$\pm$\,1.7 \\
     & ContractNLI        & 64.4\,$\pm$\,2.2 & 63.0\,$\pm$\,0.8 & 62.7\,$\pm$\,1.4 \\
     & Consumer QA        & 86.0\,$\pm$\,0.4 & 84.1\,$\pm$\,1.4 & 83.4\,$\pm$\,0.7 \\
     & \textit{Average}   & 51.4\,$\pm$\,0.2 & 50.7\,$\pm$\,0.2 & \textbf{51.7\,$\pm$\,0.2} \\
    \cmidrule(lr){1-5}
    \multirow{8}{*}{Llama-3.1-8B}
     & LegalBench         & 40.3\,$\pm$\,0.9 & 45.0\,$\pm$\,2.6 & 44.4\,$\pm$\,2.2 \\
     & CaseHOLD           & 30.0\,$\pm$\,1.9 & 37.2\,$\pm$\,1.5 & 41.4\,$\pm$\,1.2 \\
     & SARA               & 25.2\,$\pm$\,0.1 & 24.9\,$\pm$\,0.2 & 23.7\,$\pm$\,2.5 \\
     & MMLU Law           & 47.3\,$\pm$\,1.2 & 45.7\,$\pm$\,2.4 & 47.9\,$\pm$\,0.9 \\
     & MBE                & 43.6\,$\pm$\,2.2 & 50.2\,$\pm$\,0.6 & 51.7\,$\pm$\,0.7 \\
     & ContractNLI        & 58.8\,$\pm$\,2.5 & 51.1\,$\pm$\,2.4 & 62.9\,$\pm$\,1.7 \\
     & Consumer QA        & 83.1\,$\pm$\,2.1 & 83.7\,$\pm$\,2.1 & 85.0\,$\pm$\,2.8 \\
     & \textit{Average}   & 46.9\,$\pm$\,0.7 & 48.2\,$\pm$\,0.8 & \textbf{51.0\,$\pm$\,0.8} \\
    \bottomrule
    \end{tabular}
  \end{table*}

\subsection{Benchmarks}
\label{sec:benchmarks}

We evaluate the fine-tuned models on 13 held-out benchmarks, seven legal and six finance, assembled around three requirements. First, every benchmark is disjoint from the rubric datasets used for diagnosis, so no evaluation prompt influences tree construction, node scoring, or weak-node selection. Second, professional competence is not a single skill, so each suite is chosen to cover the distinct capability families its domain requires (doctrinal and factual knowledge, rule application to new facts, document-grounded interpretation, numerical reasoning, and domain-language classification) rather than many variants of one task type. Third, none of the benchmarks share the long-form format of the diagnosis data: they use multiple-choice, classification, extraction, and numeric-answer formats, so gains cannot come from memorizing a response style and must instead reflect transfer of the underlying capabilities.

The legal suite covers this spectrum as follows: \textit{MMLU Law}~\cite{mmlulawhendrycks2021} and \textit{MBE}~\cite{mbefernandes2025llamawalksbarefficient} test exam-style doctrinal knowledge; \textit{CaseHOLD}~\cite{caseholdzheng2021} tests precedent reasoning through case-holding selection; \textit{SARA}~\cite{SARAnils2020} tests statutory application to new fact patterns; \textit{ContractNLI}~\cite{koreeda-manning-2021-contractnli-dataset} tests fine-grained document-grounded contract interpretation; \textit{Consumer QA}~\cite{consumerqakolt2022predicting} tests reasoning over consumer-contract clauses; and \textit{LegalBench}~\cite{guha2023legalbench} aggregates heterogeneous legal-reasoning subtasks for breadth. The finance suite spans document-grounded QA over filings (\textit{FinanceBench}~\cite{islam2023financebench}), multi-turn arithmetic over financial reports (\textit{ConvFinQA}~\cite{chen2022convfinqa}), hybrid table-and-text numerical reasoning (\textit{TAT-QA}~\cite{zhu2021tatqa}), and three classification tasks probing understanding of market language: sentiment (\textit{FiQA-SA}~\cite{fiqa-sa-maia201818}), monetary-policy stance (\textit{FOMC}~\cite{fomc-shah2023trillion}), and ESG topics (\textit{MLESG}~\cite{mlesg-chen-etal-2023-multi-lingual}). We report token-overlap F1 for \textit{TAT-QA} because exact-match scoring under-credits verbose but partially correct reasoning traces.

This breadth makes the evaluation conservative. Several benchmarks measure capability families, such as exam-style recall and sentiment classification, that are not obviously aligned with the diagnostic data or generated training examples. Reporting the full suite therefore tests whether CRAFT improves domain-wide transfer, rather than performance on a subset close to the source data.

\subsection{Legal Benchmark Results}

Table~\ref{tab:legal_craft_wins} reports results on the legal benchmark suite. \textbf{CRAFT achieves the best legal average on three of the four models}. It improves the average for Qwen3-4B by 1.3 points over EvalTree and 3.1 points over Random, gives a smaller but positive gain for Gemma-3-4B, and gives the largest legal gain for Llama-3.1-8B, where it is 2.8 points above EvalTree and 4.1 points above Random. Qwen3-8B is the exception: EvalTree has the highest mean, but its advantage over CRAFT is 0.6 points and the decoding-variance bands overlap.

The benchmark-level results are heterogeneous, which is expected for a suite that mixes broad legal reasoning, exam-style multiple choice, statutory application, and document-grounded interpretation. CRAFT is strongest on five of seven legal benchmarks for Llama-3.1-8B and leads the domain average for Qwen3-4B and Gemma-3-4B despite baselines winning several individual rows. The legal result is therefore best read as aggregate transfer across a diverse legal suite, not as a claim that one data-selection method dominates every legal task.

  \begin{table*}[t]
    \centering
    \caption{Accuracy on finance benchmarks for Random, EvalTree, and CRAFT. Cells report mean $\pm$ standard deviation over repeated temperature-0.5 decoding runs; \textit{Average} is the unweighted benchmark mean. CRAFT performs best for all four models.}
    \label{tab:finance_craft_wins}
    \small
    \begin{tabular}{l l c c c}
    \toprule
    Model & Benchmark & Random & EvalTree & CRAFT \\
    \midrule
    \multirow{7}{*}{Qwen3-4B}
     & FinanceBench       & 6.2\,$\pm$\,1.0 & 11.6\,$\pm$\,1.0 & 12.9\,$\pm$\,1.7 \\
     & ConvFinQA          & 42.2\,$\pm$\,2.4 & 41.1\,$\pm$\,3.0 & 46.5\,$\pm$\,0.4 \\
     & TAT-QA (F1)        & 35.0\,$\pm$\,1.6 & 38.1\,$\pm$\,0.7 & 49.9\,$\pm$\,2.3 \\
     & FiQA-SA            & 63.8\,$\pm$\,1.5 & 67.5\,$\pm$\,0.5 & 70.9\,$\pm$\,2.9 \\
     & FOMC               & 50.9\,$\pm$\,2.3 & 51.0\,$\pm$\,2.0 & 58.2\,$\pm$\,3.3 \\
     & MLESG              & 27.7\,$\pm$\,0.9 & 35.3\,$\pm$\,2.0 & 37.6\,$\pm$\,0.5 \\
     & \textit{Average}   & 37.6\,$\pm$\,0.3 & 40.8\,$\pm$\,1.1 & \textbf{46.0\,$\pm$\,1.4} \\
    \cmidrule(lr){1-5}
    \multirow{7}{*}{Qwen3-8B}
     & FinanceBench       & 21.8\,$\pm$\,3.0 & 21.0\,$\pm$\,3.3 & 12.7\,$\pm$\,3.3 \\
     & ConvFinQA          & 32.3\,$\pm$\,1.3 & 32.2\,$\pm$\,0.6 & 35.1\,$\pm$\,2.3 \\
     & TAT-QA (F1)        & 48.8\,$\pm$\,1.4 & 46.5\,$\pm$\,2.2 & 36.5\,$\pm$\,0.7 \\
     & FiQA-SA            & 71.8\,$\pm$\,2.2 & 69.4\,$\pm$\,1.5 & 82.1\,$\pm$\,0.7 \\
     & FOMC               & 52.1\,$\pm$\,1.9 & 50.6\,$\pm$\,0.8 & 59.9\,$\pm$\,6.4 \\
     & MLESG              & 38.0\,$\pm$\,2.4 & 39.7\,$\pm$\,0.9 & 44.0\,$\pm$\,0.7 \\
     & \textit{Average}   & 44.1\,$\pm$\,1.1 & 43.5\,$\pm$\,0.2 & \textbf{45.1\,$\pm$\,1.9} \\
    \cmidrule(lr){1-5}
    \multirow{7}{*}{Gemma-3-4B}
     & FinanceBench       & 5.7\,$\pm$\,0.5 & 6.0\,$\pm$\,0.0 & 9.3\,$\pm$\,0.0 \\
     & ConvFinQA          & 25.1\,$\pm$\,0.0 & 25.2\,$\pm$\,1.3 & 27.0\,$\pm$\,0.8 \\
     & TAT-QA (F1)        & 32.9\,$\pm$\,0.0 & 35.0\,$\pm$\,0.6 & 37.6\,$\pm$\,0.9 \\
     & FiQA-SA            & 74.8\,$\pm$\,0.7 & 74.6\,$\pm$\,1.1 & 73.3\,$\pm$\,1.3 \\
     & FOMC               & 55.9\,$\pm$\,5.3 & 58.2\,$\pm$\,4.3 & 60.4\,$\pm$\,3.8 \\
     & MLESG              & 33.6\,$\pm$\,3.0 & 31.6\,$\pm$\,1.7 & 31.2\,$\pm$\,4.4 \\
     & \textit{Average}   & 38.4\,$\pm$\,0.5 & 38.6\,$\pm$\,0.0 & \textbf{39.8\,$\pm$\,1.4} \\
    \cmidrule(lr){1-5}
    \multirow{7}{*}{Llama-3.1-8B}
     & FinanceBench       & 9.7\,$\pm$\,0.5 & 8.7\,$\pm$\,0.0 & 7.3\,$\pm$\,1.9 \\
     & ConvFinQA          & 23.3\,$\pm$\,1.9 & 22.7\,$\pm$\,0.5 & 31.1\,$\pm$\,0.4 \\
     & TAT-QA (F1)        & 30.7\,$\pm$\,0.7 & 31.8\,$\pm$\,0.2 & 38.3\,$\pm$\,0.6 \\
     & FiQA-SA            & 81.8\,$\pm$\,2.0 & 83.1\,$\pm$\,2.3 & 83.5\,$\pm$\,1.6 \\
     & FOMC               & 48.4\,$\pm$\,2.6 & 42.5\,$\pm$\,3.6 & 57.4\,$\pm$\,2.3 \\
     & MLESG              & 26.1\,$\pm$\,3.9 & 28.9\,$\pm$\,1.6 & 35.1\,$\pm$\,1.3 \\
     & \textit{Average}   & 36.6\,$\pm$\,0.2 & 36.0\,$\pm$\,0.0 & \textbf{42.5\,$\pm$\,0.8} \\
    \bottomrule
    \end{tabular}
  \end{table*}

\subsection{Finance Benchmark Results}

Table~\ref{tab:finance_craft_wins} shows the clearest aggregate result: \textbf{CRAFT has the best
average score for every model}. The margins are largest for Qwen3-4B and Llama-3.1-8B, where CRAFT is at least five points above both baselines on the domain average. Qwen3-8B and Gemma-3-4B show smaller but still positive average gains, despite some individual benchmarks favoring a baseline.

The per-benchmark pattern is mixed but informative. CRAFT is strongest on all six finance benchmarks for Qwen3-4B and on four of six for Llama-3.1-8B; for Qwen3-8B, the average gain comes from large improvements on \textit{FiQA-SA}, \textit{FOMC}, and \textit{MLESG} that offset weaker \textit{FinanceBench} and \textit{TAT-QA} scores. The open-ended numerical benchmarks remain noisier because answers depend on long excerpts, calculations, and judge-based scoring \cite{liu-etal-2023-g, kim2024prometheus, zeng2024evaluating}. The main result is therefore the repeated domain-average pattern rather than uniform dominance on every finance dataset.

\subsection{Overall Results}

We test CRAFT on 4 open source models on 2 domains and find that it results in best downstream finetuning performance. In finance domain, CRAFT has the strongest average domain performance for all four models. In legal domain, it has the strongest domain average for three of four models; and within standard-deviation band overlap for the fourth model. The two baselines decompose this gain: CRAFT's advantage over Random, which draws from the same domain pool as a flat collection with no hierarchy, shows that organizing capabilities into a hierarchy tree and targeting its weak nodes adds value beyond untargeted in-domain data. CRAFT's advantage over EvalTree, which also builds a hierarchy but over whole prompts, shows that rubric criteria are the more informative unit to organize. More broadly, the results support the paper's central motivation: evaluation is more useful for post-training when it explains why a model fails and turns that diagnosis into targeted data for finetuning.

  \begin{table}[t]
    \centering
    \caption{This table shows ablation results for weak code selection for CRAFT. Average domain accuracy of finetuned models for fixed tree levels (L4/L5) versus CRAFT's dynamic selection shows that dynamic selection helps in identifying better weak capabilities in models.}
    \label{tab:ablation_selection}
    \small
    \setlength{\tabcolsep}{3.5pt}
    \begin{tabular}{l l c c c}
    \toprule
    Model & Domain & Fixed-L4 & Fixed-L5 & Dynamic \\
    \midrule
    Qwen3-4B     & Legal   & 47.8 & 49.0 & \textbf{50.4} \\
    Qwen3-4B     & Finance & 43.6 & 42.6 & \textbf{47.1} \\
    Llama-3.1-8B & Legal   & 43.6 & 40.1 & \textbf{50.3} \\
    Llama-3.1-8B & Finance & 42.3 & 42.9 & \textbf{43.1} \\
    \bottomrule
    \end{tabular}
  \end{table}

\subsection{Ablation: Dynamic vs. Fixed-Level Node Identification}
\label{sec:ablation}

A central design choice in CRAFT is how weak capabilities are selected from the annotated capability tree. Rather than fixing a single level of the tree and selecting the lowest-performing nodes at that level, CRAFT uses a dynamic search that surfaces underperforming nodes at whichever granularity they naturally emerge. To verify that this dynamic procedure contributes to CRAFT's advantage, and is not an interchangeable implementation detail, we run an ablation that isolates the node-selection strategy while holding everything else fixed.

We instantiate fixed-level selection at two depths of the tree: L5, the finest level, and L4, one level coarser, so that the comparison is not tied to one particular choice of granularity. Each fixed-level variant selects the same number of weak nodes as the dynamic procedure, and every other component of the pipeline is unchanged. This isolates whether dynamic selection improves performance beyond simply choosing weak nodes from the tree. We run the ablation on two models, Qwen3-4B and Llama-3.1-8B, in both the legal and finance domains, and evaluate every resulting fine-tuned model on the same benchmark suites used in the main experiments. Table~\ref{tab:ablation_selection} reports the average accuracy over each domain.

Dynamic selection achieves the best average in every ablation row. The important pattern is the instability of the fixed-depth alternatives: L4 is better in some settings, while L5 is better in others. This is exactly the failure mode CRAFT is designed to avoid. If the model's weaknesses live at different depths in the capability tree, a fixed level either fragments broad weaknesses into sparse nodes or merges narrow weaknesses into overly general ones. Dynamic selection sidesteps this choice by selecting weak nodes at the granularity where the evidence is strongest, which explains why it remains the most reliable strategy across models and domains.

\section{Conclusion}

CRAFT reframes rubric-based evaluation around a practical post-training question: what should we train on next? Standard benchmark summaries can identify weak tasks or categories, but they rarely specify which answer requirements the model failed to satisfy. CRAFT uses each rubric criterion as a capability probe, clusters prompt--rubric pairs into a scored hierarchy, and selects weak nodes dynamically across depths. The output is not a new fine-tuning recipe; it is a model-specific set of training targets derived from measured rubric failures.

Empirically, those targets matter, though not as uniform wins on every benchmark. Holding teacher generation, data budget, fine-tuning hyperparameters, and final evaluation fixed, CRAFT gives the strongest repeated-decoding finance-domain average for every model. In legal, it gives the strongest repeated-decoding domain average for three of four models and has overlapping decoding-variance bands with the best baseline on the fourth. The comparisons isolate where the gain comes from: Random shows that structure and weakness targeting help beyond additional in-domain data, EvalTree shows that rubric criteria are a more useful diagnostic unit than whole prompts in this setting. CRAFT shows strongest average performance in seven of eight model-domain settings and supports the central claim: evaluations are more useful when they do not stop at reporting low scores, and instead translate measured failures into targeted fine-tuning data.

\bibliographystyle{abbrvnat}
\bibliography{latex/custom}

\end{document}